\pdfoutput=1

\documentclass[11pt]{article}

\usepackage[final]{acl}

\usepackage{times}
\usepackage{latexsym}

\usepackage[T1]{fontenc}

\usepackage[utf8]{inputenc}

\usepackage{microtype}

\usepackage{float}
\usepackage{hyperref}
\usepackage{xurl}
\usepackage{cleveref}
\usepackage{xcolor}
\usepackage{graphicx}
\usepackage{todonotes}[inline]

\title{Great Power, Great Responsibility: Recommendations for Reducing Energy for Training Language Models}

\author{Joseph McDonald\textsuperscript{1},
  Baolin Li\textsuperscript{2},
  Nathan Frey\textsuperscript{1},
  Devesh Tiwari\textsuperscript{2},
  Vijay Gadepally\textsuperscript{1},
  Siddharth Samsi\textsuperscript{1}\\
  \textsuperscript{1}MIT Lincoln Laboratory\\
  \textsuperscript{2}Northeastern University \\
  \texttt{\{jpmcd,ncfrey\}@mit.edu}\\
  \texttt{\{li.baol,d.tiwari\}@northeastern.edu} \\
  \texttt{\{vijayg,sid\}@ll.mit.edu}
  }

\date{}

\begin{document}

\maketitle
\begin{abstract}

The energy requirements of current natural language processing models continue to grow at a rapid, unsustainable pace. Recent works highlighting this problem conclude there is an urgent need for methods that reduce the energy needs of NLP and machine learning more broadly.
In this article, we investigate techniques that can be used to reduce the energy consumption of common NLP applications. In particular, we focus on techniques to measure energy usage and different hardware and datacenter-oriented settings that can be tuned to reduce energy consumption for training and inference for language models. We characterize the impact of these settings on metrics such as computational performance and energy consumption through experiments conducted on a high performance computing system as well as popular cloud computing platforms. These techniques can lead to significant reduction in energy consumption when training language models or their use for inference. For example, power-capping, which limits the maximum power a GPU can consume, can enable a 15\% decrease in energy usage with marginal increase in overall computation time when training a transformer-based language model.\footnote{This material is based upon work supported by the Assistant Secretary of Defense for Research and Engineering under Air Force Contract No. FA8702-15-D-0001, and United States Air Force Research Laboratory Cooperative Agreement Number FA8750-19-2-1000. Any opinions, findings, conclusions or recommendations expressed in this material are those of the author(s) and do not necessarily reflect the views of the Assistant Secretary of Defense for Research and Engineering, or the United States Air Force. The U.S. Government is authorized to reproduce and distribute reprints for Government purposes notwithstanding any copyright notation herein.}




\end{abstract}


\section{Introduction}

Artificial intelligence and machine learning (ML) are increasingly used in diverse areas ranging from NLP to autonomous driving. Broadly, larger and deeper models are found to be more accurate.
However, as models 
and datasets increase in size, the computational demands of AI/ML have increased correspondingly~\citep{openai,thompson}.
In particular \citet{thompson} estimates that achieving a 10-fold improvement in model performance comes at a cost of at least a 10,000-fold increase in computation and a corresponding increase in the energy required to perform these computations.
The growth in computational and energy requirements is particularly glaring in NLP with the introduction of transformer-based language models \cite{transformer,bert,gpt}.
For example, training GPT-3 is estimated to consume almost 1300MWh \citep{berkeleygoogle}.
Given the considerable compute requirements and the associated carbon footprint of training models with increasing accuracy, there is growing interest and research into the energy demands and carbon footprint of AI~\cite{berkeleygoogle,forbes}.

However, estimating energy usage for a particular AI application depends on a number of parameters such as model architecture, hardware details, environmental parameters, and implementation details. For this reason, important works such as \citet{berkeleygoogle} and \citet{strubellacl} rely on estimates extrapolated from industry averages for some language models in some of their analysis of large neural net training, including total floating point operations per second (FLOPS) and hardware estimates, rather than values that otherwise might have been measured. Further, these articles
call for new research that identifies mitigation techniques
that can reduce the energy usage of NLP applications.

This paper proposes and characterizes potential ways to reduce the energy impact of NLP applications.
To our knowledge this is the first presentation of power-capping as a useful tool for reducing GPU energy consumption.
Particularly in the context of deep learning and NLP, this work provides an approach alongside estimates for possible energy savings for training large, energy-intensive language models.
Moreover this method does not affect the predictions of trained models or consequently their performance accuracy on tasks.
That is, if two networks with the same structure, initial values and batched data are trained for the same number of batches under different power-caps, their resulting parameters will be identical and only the energy required to produce them may differ.
Section 2 presents related work on tracking energy usage of NLP applications and their environmental impact.
Section 3 introduces different techniques, including power-capping and energy-aware scheduling, that can be used to reduce the energy usage, including experiments and other relevant data to characterize their effectiveness.
In Section 4, we discuss these approaches with broader recommendations before concluding with future avenues of research.



\section{Prior work}

Energy efficiency considerations for deep learning have trailed model developments targeted at improving accuracy among other metrics with new, often growing architectures.
Highlighting this focus, the growth of neural network architecture sizes is considered in \citet{canziani}. That study offers a comparison of state-of-the-art image recognition models where their computational performance is analyzed including inference time and power utilization.
Techniques for model compression have been widely studied including knowledge distillation and pruning \citep{distillation,lottery}.
In NLP, distillation has been used to reduce the size of large language models \cite{distilbert}, and other methods of compression have been effective at shrinking model parameters such as embedding layers \cite{mu2018,compressw2v}.



Recent attention has focused on the size of NLP models alongside their extensive training times and environmental impact \cite{strubellacl,berkeleygoogle,allenai}.
These works illustrate efforts to place greater consideration on the efficiency or inefficiencies of large neural network architectures.
For instance, \citet{allenai} weighs the advantages of different metrics to evaluate efficiency while advocating for the use of floating point operations as a way to objectively compare models.
Another area of focus has been on the dependence of model test accuracies on the amount of computation expended on hyperparameter tuning \cite{show}.
Some of these works propose considering efficiency alongside accuracy as a metric for evaluating ML models, and at the very least to require reporting energy consumption and carbon impact used in research for conference and journal submissions.

While calls to prioritize more efficient methods of training NLP models are made in the previously cited papers
among other works, to the best of our knowledge this is not  reflected in publicly available academic or industry  research.
In fact, in \citet{henderson} a random sampling of 100 NeurIPS papers showed that few papers tracked and reported these statistics -- and none reported carbon impact.
This and the previous works point out that tracking energy usage is not yet a standard practice, in part because of the difficulty in implementing a framework for collecting these statistics from hardware.
An implementation for accurately capturing this data on common hardware (specifically Intel and NVIDIA hardware) is presented in \citet{henderson} which relies on querying device software tools.
We describe another, similar approach for gathering power expenditure and energy usage in this work, in order to present a straightforward process for obtaining accurate measurements of energy consumption.



Compared to these works, this paper presents steps that can be taken to reduce the energy required for training and inference with language models.
There is limited prior research investigating power-capping as a method for reducing energy consumption \cite{powercapcpu}, and it has focused on CPUs for scientific computing applications.
Our focus is specifically on widely-used AI/ML frameworks used with available commodity hardware.
This approach is described with experiments showcasing its effectiveness for a range of settings. 
Additionally our findings for GPU energy reduction when training neural networks are comparable and consistent with the outcomes for CPU consumption presented in \citet{powercapcpu}.

Similar recent work investigates distributed DNN training fitting power law models that describe how training time scales with available compute resources and energy constraints \citep{dnnscaling}.
Additionally we address other approaches towards reducing energy footprints by considering shifting habits in training.
Utilizing datacenters and climate-aware workload scheduling can provide considerable savings, and we share statistics from our institutional datacenter to support this \citep{supercloud,scdataset}.

\section{Reducing the Energy Impact of NLP}\label{sec:reducing}

This section outlines various approaches that can be used to reduce the energy consumption of NLP workloads.
We focus primarily on a simple yet effective method -- power-capping -- that yields significant benefit with minimal cost and translates across different computing platforms.
Experiments measuring the effect of power-caps on energy consumption are presented.
For completeness, we discuss other potential avenues for reducing the carbon impact of NLP applications.
Data is presented for the monthly and daily variation in energy efficiency of our institution's datacenter. This illustrates in detail how much energy usage can be reduced by simple approaches like timing workloads to certain hours or seasons if possible.
While factors like efficiency and daily variation depend heavily on characteristics unique to each organization's datacenter, we share general insights that will hold true for most cases.

\noindent \textbf{Measuring Energy Usage:} Currently, there are two vendor-provided utilities to monitor resource consumption on NVIDIA GPUs. The NVIDIA Data Center GPU Manager (DCGM) is a suite of tools for managing and monitoring NVIDIA GPUs in cluster environments~\cite{dcgm} and the NVIDIA System Management Interface~\cite{nvidia-smi} (or \texttt{nvidia-smi}) utility, which can also perform similar monitoring.  Broadly, these tools enable monitoring of GPU usage on a node and the collection of metrics on Streaming Multi-processor (SM) utilization, GPU memory footprint, power draw, GPU temperatures, PCI Express (PCIe) bandwidth, and several other hardware settings. On our system, this data is collected on every node and every GPU assigned to a job. The data is collected every 100ms and data collection is started and stopped automatically using the scheduler that manages resources on the system.

\subsection{Limiting Hardware Power}
\begin{figure}[b!]
    \centering
    \includegraphics[width=.5\textwidth]{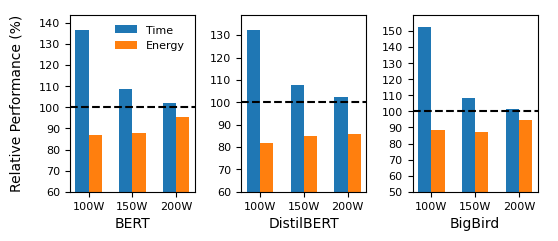}
    \caption{\label{fig:plotmodels} Time and energy usage comparison of training three language modeling network architectures with different maximum power limits. Values given are percentage relative to performance of default 250W setting (100\% indicated by black line). For example, training BERT with a 150W limit required 108.5\% of the time and only 87.7\% the energy needed to train with default settings.
    }
\end{figure}
\begin{figure*}[t!]
    \centering
    \includegraphics[width=\textwidth]{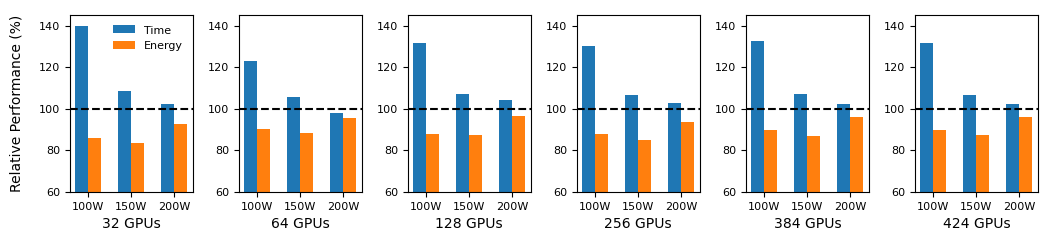}
    \caption{\label{fig:plotscaling} Time and energy required for training with varying number of GPUs at different power thresholds. Values are the percent relative to time or energy required for the default setting of 250W.
    Average relative time for 150W is indicated by blue line, and average relative energy consumption for 150W is indicated by orange line. For 32, 64, 128, 256, 384 and 424 GPUs, training was performed for 6, 10, 15, 25, 40, and 40 epochs respectively to ensure similar job durations.
    In most cases, power-capping required additional time to complete training but resulted in less overall energy consumption.
}
\end{figure*}

Most modern computing platforms allow users to adjust hardware settings for processors and GPUs. This can be done via command line tools that are generally not visible to users of a shared computing system. Over the duration of an NLP task, the power consumed by hardware components can vary significantly based on the operation being performed, environmental conditions, and hardware limits.
Power-capping allows users to limit the maximum power available to hardware devices through these tools.
On our cluster, this is implemented using the \texttt{nvidia-smi} command line utility.

A series of experiments is presented here that use energy tracking tools to measure the reduction in energy consumption provided by power-capping GPUs.
Power-capping requires no changes to user code and is done at a hardware level. Below, we validate these savings for various scenarios such as how these savings translate across different models for masked language modeling (MLM) or how these savings work across different sets of resources and hardware platforms. From our observations, this method provides a noticeable difference in all scenarios with very little incurred effect on computation time.
For these experiments we use a popular PyTorch implementation for MLM from Hugging Face\footnote{\url{https://github.com/huggingface/transformers/blob/master/examples/pytorch/language-modeling/run_mlm.py}}.

\noindent\textbf{Power-capping works across different models}:
We train different transformer-based networks -- BERT, DistilBERT, and Big Bird \cite{bert,distilbert,bigbird} -- with MLM and observe that power-capping is beneficial to energy usage regardless of architecture.
Each model was trained on 16 V100 GPUs using four different power caps: 100 watts (W), 150W, 200W and 250W (the default power limit for an NVIDIA V100 GPU on our system).
Models were trained with the WikiText-103 \cite{wikitext} dataset for 4 epochs and batches of 8 samples per GPU.
Network parameters were trained from scratch with randomly initialized values,
and random number seeds fixed for consistency across runs with different power thresholds.

\Cref{fig:plotmodels} depicts training performance with power-capping at 100W, 150W and 200W. Results are plotted as a percent relative to the default limit of 250W. Our experiments indicate that implementing power caps can significantly reduce energy usage at the cost of training time.


\noindent\textbf{Energy savings at larger scales:}
We performed a similar test training BERT with MLM on distributed configurations of varying numbers of GPUs.
Energy measurements were gathered for each training run on different node configurations equipped with between 2 and 400+ GPUs and the same choices for power limits as before.
Models were trained on WikiText-103 with a batch size of 8 samples per GPU.

The time and energy required for training at different power thresholds is given in \Cref{fig:plotscaling}, where values are the percent relative to time or energy required for the default setting of 250W.
Averaging across each choice of configuration, a 150W bound on power utilization led to an average 13.7\% decrease in energy usage and 6.8\% increase in training time compared to the default maximum.
Note from \Cref{fig:plotscaling} that the 100W setting has significantly longer training times (31.4\% longer on average). A 200W limit corresponds with almost the same training time as a 250W limit but more modest energy savings than a 150W limit.
These outcomes support the use of power-capping at 150W for this GPU architectures and this application.
We expect that different applications may require different settings for optimal efficiency which could be identified empirically. 

\begin{figure}[t!]
    \centering
    \includegraphics[width=.5\textwidth]{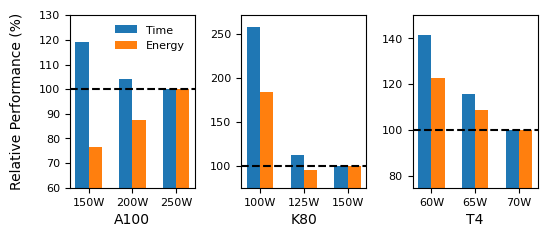}
    \caption{\label{fig:plothardware}
    Performance impact of power-caps on different NVIDIA GPUs relative to default limits, 250W on A100, 150W on K80, and 70W on T4.
    Limiting maximum power has a significant effect on each platform. For A100s the effect is similar to the V100s we test in other cases, if more pronounced with greater energy savings for both the 150W and 200W settings.
    However for T4 architectures the default 70W settings perform optimally, and the effect for K80s is less clear.
    }
\end{figure}

\noindent\textbf{Energy savings translate across hardware platforms:} We performed additional experiments across several different GPUs used widely in ML research to check this method's effectiveness. This was tested on NVIDIA's K80, T4 and A100 GPUs, available through our institution's HPC resources as well as Amazon Web Services.
\Cref{fig:plothardware} presents these results.
While there is not a single obvious choice for optimal settings, we confirm that the effect of power-capping is not limited to one type of hardware platform.
In each of the platforms, modifying the maximum power limit affected the efficiency of the device.
For A100s the effect is similar to the V100s discussed previously, if more pronounced with greater energy savings for both the 150W and 200W settings.
However for T4 processors the default 70W settings perform optimally, and the effect for K80s is less clear.
Many factors affect how much power is needed for efficient GPU computation, and memory intensive batch training required by language models as well as hardware specific behaviors could lead to poorer performance on these older NVIDIA architectures.

\noindent\textbf{Energy savings apply to inference:}
Different estimates from NVIDIA and Amazon suggest that inference tasks account for 80\% or more of AI computational demand \cite{amazoninference,nvidiainference} while training new models is responsible for a much smaller fraction.
Thus, methods for reducing energy on inference tasks can have a greater impact in reducing AI's carbon footprint compared to training.

We measure the effect of power-capping applied to hardware when performing inference with a trained BERT model.
This test was limited to a single node with two V100 GPUs as inference is naturally parallelizable across multiple devices.
Measurements show that power-capping has a more pronounced effect for inference tasks on running time and energy usage.
Compared to 250W, a 100W setting required double the inference time (a 114\% increase) and consumed 11.0\% less energy, 150W required 22.7\% more time and saved 24.2\% the energy, and 200W required 8.2\% more time with 12.0\% less energy.
For language modeling with BERT, energy gains through power-capping are noticeably greater when performing inference than for training.
If this is consistent for other AI applications, this could have significant ramifications in terms of energy consumption for large-scale or cloud computing platforms serving inference applications for research and industry.





\subsection{Energy-aware Scheduling}
AI and NLP researchers often rely on HPC datacenters managed by cloud computing providers or their institutions if available.
The efficiency of a datacenter varies through the day as well as through the year. A common metric used across the datacenter community to measure datacenter efficiency is Power Usage Effectiveness (PUE) defined as 
\begin{equation}
PUE=\frac{FE+IT}{IT}
\end{equation}
where $IT$ is the information technology energy and $FE$ is the facility energy.
Facility energy includes energy consumed by the datacenter to perform climate control and any additional energy required for operating the computing equipment.
IT energy includes the energy used by computing hardware.
A highly efficient datacenter will have a PUE close to 1, such that the facility energy overhead is minimal, while the global average for PUE is 1.59 \citep{pue}. A PUE of 1.59 indicates that nearly 40\% of a datacenter's energy usage is consumed by facility energy. 

\begin{figure}[t]
    \vspace{-0.1in}
    \centering
    \includegraphics[width=\columnwidth]{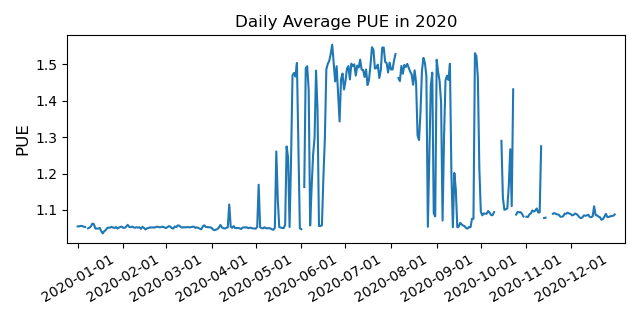}
    \caption{\label{fig:pue_year} PUE measurements averaged for each day throughout 2020. Hotter summer temperatures correspond to more energy required for cooling compute resources and greater PUE values.}
\end{figure}
\begin{figure}[t]
    \centering
    \includegraphics[width=\columnwidth]{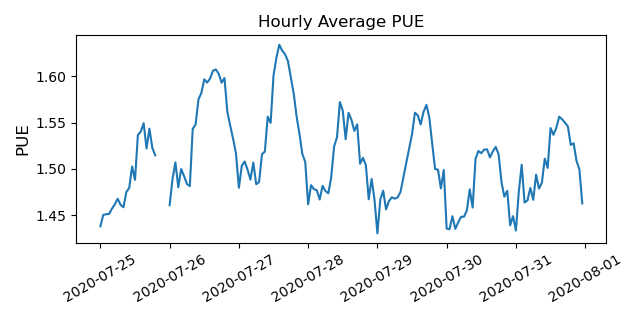}
    \caption{\label{fig:pue_week} Average hourly variation in PUE for our datacenter over one week in July 2020. Measurements tend to peak during hot afternoon hours and decrease throughout cooler night temperatures. For example the hourly minimum on July 27 is 1.48 from 12--1 a.m. while the maximum is 1.63 between 2--3 p.m., translating to a daily variation of 10.4\%.
    }
\end{figure}
\begin{table}[t]
\vspace{-0.1in}\centering
\begin{tabular}{p{0.2\linewidth}cccc} 
 \hline
Month & PUE Variation (\%)  \\
 \hline
 January & 1.30 \\ 
 February & 0.69 \\ 
 March & 0.77 \\ 
 April & 2.15 \\ 
 May & 11.51 \\ 
 June & 21.70 \\ 
 July & 7.76 \\ 
 August & 17.37 \\ 
 September & 12.41 \\ 
 October & 8.07 \\ 
 November & 2.88 \\ 
 December & 1.07 \\
 \hline
 Annual & 7.30 \\
 \hline
\end{tabular}
\caption{\label{tab:puetable} Average daily variation in PUE for each month at our institution's datacenter in 2020. A single day's PUE variation is the percent difference of the hour with the greatest average PUE and the hour with the minimum PUE average. The monthly variation is the average of this value over days in the month. The annual variation is the average over all days in the year, and not the average among the months.}
\end{table}

In \Cref{fig:pue_year} the average daily PUE measurements from our institutional datacenter are plotted for the entirety of 2020, showing how seasonal changes in temperature can affect the energy consumption of any individual computational workload.
For instance the average PUE in January is 1.05 while in July it is 1.49, a 42\% difference.
Evidently, heavy NLP workloads are typically much less efficient in the summer than those executed during winter.
Given the large seasonal variation, if there are computationally expensive experiments that can be timed to cooler months this timing can significantly reduce the carbon footprint.

To show how resource efficiency can vary even over relatively short periods of time, our datacenter's average hourly PUE across the last week of July 2020 is plotted in \Cref{fig:pue_week}.
Each point in the curve is the average of the several PUE measurements taken each hour, so that the swings in efficiency between daytime and night hours can be readily observed.
Daytime peaks result from extra energy required for cooling while outside temperatures are high.
For instance, on July 27 the PUE peaks at 2 p.m. at an average of 1.63 while ten hours later the average measurement is 1.46, a 12\% difference.

We consider the variation in PUE over the course of a day, where the variation is the percent difference of the day's maximum hourly average compared to the minimum hourly average.
The monthly variation is the average of this percent difference over every day of the month and is listed in \Cref{tab:puetable}. 
The annual variation is the average over all days in the year, not the average among the months.
Daily variation of PUE is 7.3\% on average -- with larger daily swings in the summer months and smaller swings in the winter months.


Significant energy savings can be obtained if workloads can be scheduled at times when a lower PUE is expected.
For example, moving a short-running job from daytime to nighttime may provide a roughly 10\% reduction, and moving a longer, expensive job (e.g. a language model taking weeks to complete) from summer to winter may see a 33\% reduction.
While it is difficult to predict the savings that an individual researcher may achieve, the information presented here highlights the importance of environmental factors affecting the overall energy consumed by their workloads.

\subsection{Relaxing Training Duration} \label{sec:relax}

\begin{figure}[t!]
    \centering
    \includegraphics[width=\columnwidth]{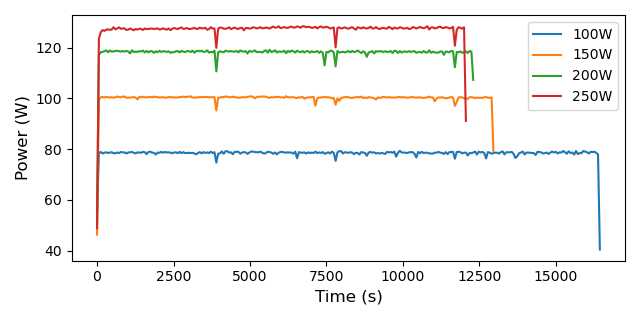}
    \caption{\label{fig:avg_power} Examples of sampled power measurements for four identical jobs at different power-cap thresholds are presented, where points in each curve give averages over one minute intervals. Note that average power remains consistent for the duration of each job.}
\end{figure}

In training different models we tracked energy consumption throughout each run and observed that the rate of energy consumption (power) is roughly constant after averaging over short intervals (one minute in this case).
This is depicted in \Cref{fig:avg_power} for four jobs with identical parameters but different power-cap limits.
It can be expected that cutting training time by $X$ percent will correspond to an $X$ percent reduction in energy.
We highlight this in consideration of common practices of significantly extending training times for marginal performance gains.
For instance in \cite{bert} doubling the number of training batches provided an additional 1\% increase in performance on a particular benchmark test set.
For certain applications or domains this additional training may make sense, but in cases where evaluation metrics include energy considerations, longer training for marginal performance improvements would be counterproductive and could incur significant energy expenditure.

\subsection{Utilizing Efficient Datacenters}

One last practice we address that can help researchers reduce their environmental impact is utilizing institutional shared datacenters and cloud computing resources for energy-intensive NLP applications.
By considering this approach for applications as opposed to building and managing smaller, private HPC workstations or clusters, researchers can save money on equipment purchases and potentially energy bills depending on where resources are housed, as well as reducing the carbon footprint of their workloads.
While there is convenience in having private computing resources that are accessible, this convenience comes at a cost.
Generally speaking energy savings and impact is more easily obtained at larger scales.
Datacenters and cloud computing providers make significant investments in the efficiency of their facilities.
For instance, Google publishes data on its PUE, reaching a 12 month average of 1.10 in 2021\footnote{\url{https://www.google.com/about/datacenters/efficiency/}}, and the National Renewable Energy Laboratory sets an annual goal of running their computing facility at a PUE of less than 1.06 \footnote{\url{https://www.nrel.gov/computational-science/measuring-efficiency-pue.html}}, recently achieving a record of 1.036.

Additionally, many cloud providers are moving their energy supply towards more environmentally friendly and renewable energy sources in attempts to reduce their carbon output to zero \cite{amazonpue}.
These types of improvements would be time-consuming and difficult to make for individual researchers,
but by sacrificing some conveniences, AI researchers can reap these benefits without additional effort beyond moving their projects to these platforms.

\section{Discussion and Recommendations}
We believe the approaches proposed here offer easy-to-implement solutions for reducing the carbon footprint of NLP applications without significant algorithmic or software changes.
Though they do not involve new algorithmic methods which are outside the scope of this article, these represent early steps towards more efficient NLP.
Coupling them with algorithmic changes would further improve energy consumption.
The goal of this article is to initiate a conversation between NLP researchers and those in the hardware and datacenter domains.
Below, we list additional recommendations that may help shape such a conversation.



\noindent\textbf{Understanding your computational environment's characteristics:}
Previous works highlighted the carbon footprint of computationally expensive NLP applications, and their recommendations of tracking and reporting energy usage was intended to encourage researchers to be aware of their individual impact.
Similarly we highlight the importance of datacenter characteristics and PUE variation to promote a deeper understanding of research energy requirements and the factors that constitute them.
We hope this work leads NLP researchers to question assumptions about the datacenters where their workloads are running and what the relative efficiency of those datacenters are. For example, researchers should opt for energy-efficient datacenters and encourage their organizations to deploy or leverage energy-efficient datacenters. 
If possible, it would also be helpful for researchers to learn these operating characteristics of their datacenters or computing providers.
Further we encourage the NLP community to work with their computing facility or datacenter to implement frameworks for tracking energy consumption like that outlined in Section \ref{sec:reducing} and other works (e.g. \citet{henderson}).

\noindent\textbf{Promoting better energy usage:} In recent years, top conferences in AI and machine learning have introduced the requirement that papers include an ethics statement addressing the potential impact of their work on the broader society. However, one area that is currently lacking is the impact of AI on the environment.
It may be difficult to account for every trial run or hyperparameter tuning when tracking and reporting estimates of energy usage.
However, we hope that this practice promotes better awareness of AI energy consumption and fosters a greater focus on optimization pathways to reduce energy usage.

Alongside reporting energy consumption statistics, we make the additional recommendation that conference papers' statements discussing ethical considerations also identify steps undertaken to minimize energy consumption. We give an example of such an ``energy statement" after concluding.
Additionally, it is critical that energy-efficient NLP research be promoted in the research community, perhaps via specialized tracks or workshops focused on these problems.


\noindent\textbf{Reducing the environmental impact:}
The purpose of this research is to
educate NLP researchers on tools that can be used to reduce their energy usage and
empower them to leverage those tools to minimize their carbon footprint.
The methods discussed fit into a wider research effort to enable more efficient AI.
Lastly, we also echo earlier calls for promoting more energy-conscious NLP practices and discourage overtraining or extensive hyperparameter searches.
Reviews of conference submissions should consider whether new methodologies are effective or the result of expensive optimization.


\section{Conclusions}

This article presents techniques that can improve the energy efficiency of training and inference for NLP applications.
Importantly, the methods discussed can be used jointly with each other to achieve a compounding effect of energy savings.
Future work relevant to these topics would include a wider survey of AI hardware and power-capping capabilities. While we focused on NVIDIA GPUs, evaluation of AI hardware from other vendors and cloud providers could have a potentially large impact for cloud computing as well as large shared high performance computing centers.  


\section*{Energy Statement}
\label{sec:energy}

The experiments performed in this work consumed a total of 782 kWh. A majority of the experiments (approximately 760 kWh) were performed on our institution's high performance computing cluster, powered by largely carbon-free, hydroelectric power sources.
To minimize energy consumption, much of these experiments were performed during system downtimes (e.g., when the system is undergoing scheduled maintenance and less busy) and when cooling needs are reduced.

\section*{Acknowledgements}
The authors acknowledge the MIT Lincoln Laboratory Supercomputing Center (LLSC) for providing HPC resources that have contributed to the research results reported in this paper. The authors wish to acknowledge the following individuals for their contributions and support: Bob Bond, Tucker Hamilton, Jeff Gottschalk, Tim Kraska, Mike Kanaan, CK Prothmann, Charles Leiserson, Dave Martinez, John Radovan, Steve Rejto, Daniela Rus, Marc Zissman, Matthew L Weiss, David Bestor, Michael Jones, Albert Reuther, William Arcand, William Bergeron, Chansup Byun, Michael Houle, Matthew Hubbell, Hayden Jananthan, Jeremy Kepner, Kurt Keville, Anna Klein, Adam Michaleas, Peter Michaleas, Lauren Milechin, Julia Mullen, Charles Yee, Andrew Prout, and Antonio Rosa. We also acknowledge support from NSF awards 1920020, 2124897, and 1910601, the Massachusetts Green High Performance Computing Center, and Northeastern University.

\bibliography{anthology,refs}
\bibliographystyle{acl_natbib}
\end{document}